\documentclass[aps,prd,twocolumn,superscriptaddress,nofootinbib]{revtex4-2}

\usepackage{amsmath,amssymb}
\usepackage{graphicx}
\usepackage{hyperref}
\usepackage{booktabs}
\usepackage{xcolor}

\hypersetup{colorlinks=true,linkcolor=blue,citecolor=blue,urlcolor=blue}

\begin{document}

\title{Spectral Signatures of Data Quality:\\Eigenvalue Tail Index as a Diagnostic\\for Label Noise in Neural Networks}

\author{Matthew Loftus}
\email{matthew.a.loftus@gmail.com}
\affiliation{Independent researcher}

\date{\today}

\begin{abstract}
We investigate whether spectral properties of neural network weight matrices can predict test accuracy. Under controlled label noise variation, the tail index $\alpha$ of the eigenvalue distribution at the network's bottleneck layer predicts test accuracy with leave-one-out $R^2 = 0.984$ (21 noise levels, 3 seeds per level), far exceeding all baselines: the best conventional metric (Frobenius norm of the optimal layer) achieves LOO $R^2 = 0.149$. This relationship holds across three architectures (MLP, CNN, ResNet-18) and two datasets (MNIST, CIFAR-10). However, under hyperparameter variation at fixed data quality (180 configurations varying width, depth, learning rate, and weight decay), all spectral and conventional measures are weak predictors ($R^2 < 0.25$), with simple baselines (global $L_2$ norm, LOO $R^2 = 0.219$) slightly outperforming spectral measures (tail $\alpha$, LOO $R^2 = 0.167$). We therefore frame the tail index as a \emph{data quality diagnostic}: a powerful detector of label corruption and training set degradation, rather than a universal generalization predictor. A noise detector calibrated on synthetic noise successfully identifies real human annotation errors in CIFAR-10N (9\% noise detected with 3\% error). We identify the information-processing bottleneck layer as the locus of this signature and connect the observations to the BBP phase transition in spiked random matrix models. We also report a negative result: the level spacing ratio $\langle r \rangle$ is uninformative for weight matrices due to Wishart universality.
\end{abstract}

\maketitle

\section{Introduction}
\label{sec:intro}

A central question in deep learning theory is: \emph{why do overparameterized neural networks generalize?} Networks with far more parameters than training examples routinely achieve low test error, contradicting classical statistical learning theory~\cite{zhang2017understanding}. Understanding when and why generalization occurs---and predicting it without a held-out test set---remains an open problem.

Random matrix theory (RMT) offers a natural framework for studying weight matrices. At initialization, weight matrices are random, and their spectral properties are well-characterized by the Marchenko-Pastur (MP) law~\cite{marchenko1967distribution}. During training, the spectrum evolves away from MP as the network learns structure from data. Martin and Mahoney~\cite{martin2021implicit} showed that well-trained networks develop heavy-tailed eigenvalue distributions and proposed the tail index $\alpha$ as a quality metric. Subsequent work demonstrated that $\alpha$ predicts model \emph{quality} (with a bias toward CV models) rather than \emph{generalization} in the universal sense~\cite{martin2020predicting}, with similar findings in NLP~\cite{yang2023evaluating}.

In this work, we conduct a controlled study of spectral predictors under two distinct perturbation axes: (1) label noise variation at fixed architecture, and (2) hyperparameter variation at fixed data quality. This design reveals that spectral measures are powerful diagnostics for data quality degradation but not universal generalization predictors. We present both positive and negative results honestly.

\subsection{Summary of contributions}

\begin{enumerate}
\item \textbf{Strong data quality prediction:} The tail index $\alpha$ of the bottleneck layer predicts test accuracy with LOO $R^2 = 0.984$ across 21 label noise levels with 3 seeds each, far exceeding all baselines ($R^2 \leq 0.149$) (Sec.~\ref{sec:prediction}).

\item \textbf{Honest scope:} Under hyperparameter variation at fixed data quality, all measures (spectral and conventional) are weak predictors ($R^2 < 0.25$), and simple baselines slightly outperform spectral measures (Sec.~\ref{sec:hyperparams}).

\item \textbf{Bottleneck hypothesis:} The spectral signature concentrates at the information-processing bottleneck---the layer with the highest compression ratio and sufficient spectral resolution (Sec.~\ref{sec:bottleneck}).

\item \textbf{Comprehensive baselines:} We compare tail $\alpha$ against effective rank, Frobenius norms, spectral norms, and global $L_2$ norm, establishing clear relative strengths (Sec.~\ref{sec:baselines}).

\item \textbf{Methodological correction:} The level spacing ratio $\langle r \rangle$ is uninformative for rectangular weight matrices due to Wishart universality (Sec.~\ref{sec:level_spacing}).

\item \textbf{Theoretical framework:} We connect our observations to the BBP phase transition and derive an informal bound on outlier eigenvalue count (Sec.~\ref{sec:theory}).
\end{enumerate}

\section{Background}
\label{sec:background}

\subsection{Marchenko-Pastur law}

For a random matrix $W \in \mathbb{R}^{m \times n}$ with iid entries of variance $\sigma^2$, the eigenvalues of the Gram matrix $S = \frac{1}{m}W^\top W$ follow the Marchenko-Pastur distribution with aspect ratio $\gamma = n/m$ and support $[\lambda_-, \lambda_+]$ where
\begin{equation}
\lambda_\pm = \sigma^2(1 \pm \sqrt{\gamma})^2.
\end{equation}
Eigenvalues outside this support indicate learned structure.

\subsection{Level spacing ratio}

The level spacing ratio $\langle r \rangle$, defined as $r_i = \min(s_i, s_{i+1})/\max(s_i, s_{i+1})$ where $s_i = \lambda_{i+1} - \lambda_i$, distinguishes uncorrelated (Poisson, $\langle r \rangle \approx 0.386$) from correlated (GOE, $\langle r \rangle \approx 0.531$) eigenvalue statistics~\cite{oganesyan2007localization}.

\subsection{BBP transition}

The Baik-Ben Arous-P\'ech\'e (BBP) transition~\cite{baik2005phase} shows that a rank-1 perturbation of strength $\theta$ to a Wishart matrix produces an outlier eigenvalue above $\lambda_+$ if and only if $\theta > \sigma^2\sqrt{\gamma}$. This provides a sharp threshold for when learned structure becomes spectrally visible.

\subsection{Heavy-tailed self-regularization}

Martin and Mahoney~\cite{martin2021implicit} observed that trained networks develop heavy-tailed eigenvalue distributions, with the tail index $\alpha$ (estimated via the Hill estimator~\cite{hill1975simple}) serving as a quality metric. They reported correlations between $\alpha$ and reported test accuracy across published models, but did not establish a quantitative predictive relationship under controlled conditions.

\subsection{Generalization measures}

A number of norm-based generalization measures have been proposed, including path norms~\cite{neyshabur2015norm}, spectral norms with margin~\cite{bartlett2017spectrally}, and PAC-Bayes bounds~\cite{dziugaite2017computing}. Jiang et al.~\cite{jiang2020fantastic} conducted a large-scale comparison of generalization measures, finding that no single measure dominates across all perturbation types. Our work complements this by studying spectral measures specifically and identifying the perturbation axis (data quality vs.\ hyperparameters) as the key determinant of predictive power.

\section{Methods}
\label{sec:methods}

\subsection{Spectral observables}

For each weight matrix $W \in \mathbb{R}^{m \times n}$, we compute the eigenvalues $\{\lambda_i\}$ of $S = \frac{1}{m}W^\top W$ and measure:

\begin{itemize}
\item \textbf{Tail index} $\alpha$: Hill estimator~\cite{hill1975simple} on eigenvalues above the 90th percentile. Lower $\alpha$ indicates heavier tails (more concentrated signal).
\item \textbf{Effective rank} (normalized): $\exp(H(\mathbf{p}))/n$ where $H(\mathbf{p}) = -\sum p_i \log p_i$ and $p_i = \lambda_i / \sum \lambda_j$. Values near 1 indicate uniform spectrum; values near 0 indicate rank-deficient.
\item \textbf{Outlier fraction}: Fraction of eigenvalues above $\lambda_+ = \sigma^2(1+\sqrt{\gamma})^2$, where $\sigma^2$ is estimated from the initial (untrained) weight matrix.
\item \textbf{MP deviation}: Kolmogorov-Smirnov statistic between the empirical eigenvalue distribution and the best-fit MP distribution.
\end{itemize}

\subsection{Baseline measures}
\label{sec:baselines}

We compare spectral observables against conventional norm-based measures:

\begin{itemize}
\item \textbf{Global $L_2$ norm}: $\sqrt{\sum_l \|W_l\|_F^2}$, the total parameter norm.
\item \textbf{Sum of Frobenius norms}: $\sum_l \|W_l\|_F$.
\item \textbf{Best-layer Frobenius norm}: $\max_l |\text{corr}(\|W_l\|_F, \text{test\_acc})|$, choosing the single layer whose Frobenius norm best correlates with test accuracy.
\item \textbf{Max spectral norm}: $\max_l \sigma_1(W_l)$, the largest singular value across all layers.
\item \textbf{Product of spectral norms}: $\prod_l \sigma_1(W_l)$.
\end{itemize}

\subsection{Architectures}

We test three architectures:
\begin{itemize}
\item \textbf{MLP:} $[784, 512, 256, 128, 10]$ on MNIST, SGD with momentum 0.9, lr=0.01, 30 epochs.
\item \textbf{CNN:} 4 conv layers (64, 64, 128, 128 channels) + 2 FC layers (256, 10) on CIFAR-10, SGD, lr=0.01, 50 epochs, no weight decay or data augmentation.
\item \textbf{ResNet-18:} Modified for 32$\times$32 input (3$\times$3 first conv, no maxpool) on CIFAR-10, SGD, lr=0.01, 50 epochs.
\end{itemize}

\subsection{Label noise protocol}
\label{sec:noise_protocol}

We use two experimental designs:

\textbf{EXP-010 (label noise gradient):} For the MLP/MNIST architecture, we train with 21 evenly spaced label noise fractions $\eta \in [0, 1]$ (i.e., $\eta \in \{0, 0.05, 0.10, \ldots, 1.0\}$), with 3 independent random seeds per noise level (63 total training runs). A fraction $\eta$ of training labels are replaced with uniformly random labels. Test labels are always clean. This produces a fine-grained gradient from full generalization ($\eta=0$) to pure memorization ($\eta=1$). All metrics are evaluated using leave-one-out (LOO) cross-validation across the 21 noise levels, with seed-averaged values and error bars.

\textbf{EXP-011 (hyperparameter variation):} For MLP/MNIST with \emph{clean labels} ($\eta=0$), we train 180 configurations (3 seeds each, 540 total runs) varying width $\in \{64, 128, 256, 512\}$, depth $\in \{2, 3, 4\}$, learning rate $\in \{0.001, 0.01, 0.1\}$, and weight decay $\in \{0, 10^{-4}, 10^{-3}, 10^{-2}, 10^{-1}\}$.

\subsection{Evaluation metric}

All predictive comparisons use leave-one-out (LOO) cross-validated $R^2$, which penalizes overfitting and provides an honest estimate of out-of-sample predictive power. This is a deliberately conservative metric: each prediction is made without access to the corresponding data point. Our earlier experiments with 6 noise levels gave inflated $R^2$ values ($>0.97$) due to the small sample size; the 21-level LOO $R^2$ reported here is the definitive result.

\subsection{Null model}

As a control, we compare spectral properties of networks trained on real labels versus fully shuffled labels ($\eta=1$), matched for training duration and convergence level.

\section{Level Spacing is Uninformative for Weight Matrices}
\label{sec:level_spacing}

Our initial hypothesis was that $\langle r \rangle$ would transition from Poisson (at initialization) to GOE (after training), with the transition speed or final value distinguishing generalization from memorization.

\textbf{This hypothesis is incorrect.} We found that $\langle r \rangle \approx 0.53$ (GOE) at initialization and remains at GOE throughout training, for both real and shuffled labels (Fig.~\ref{fig:bulk_comparison}, bottom-right panel).

The explanation is Wishart universality: for any random matrix $W$ with iid entries (regardless of distribution), the Gram matrix $\frac{1}{m}W^\top W$ follows the Wishart-Laguerre ensemble, which has GOE-like level spacing by construction. Since weight matrices are initialized with iid entries and SGD perturbations are small relative to the total matrix norm, $\langle r \rangle$ remains at GOE throughout training.

\textbf{Lesson:} The level spacing ratio, designed for Hamiltonian matrices in quantum chaos, is the wrong observable for rectangular weight matrices. Bulk distribution properties (tail index, outlier fraction, effective rank) are the correct tools.

\section{Bulk Observables Distinguish Generalization from Memorization}
\label{sec:bulk}

Having identified the correct observables, we compare networks trained on real labels versus shuffled labels (MLP/MNIST).

At convergence (both achieving $>99\%$ training accuracy), the spectral signatures differ dramatically (Table~\ref{tab:bulk}).

\begin{table}[h]
\caption{Spectral properties of MLP input layer (784$\to$512) at convergence. Real labels: 98.3\% test accuracy. Shuffled labels: 9.9\% test accuracy.}
\label{tab:bulk}
\begin{tabular}{lccc}
\toprule
Observable & Init & Real & Shuffled \\
\midrule
Outlier fraction & 0\% & 6.4\% & 76\% \\
Tail $\alpha$ & 6.5 & 2.1 & 3.5 \\
Effective rank (norm) & 0.72 & 0.61 & 0.47 \\
MP KS & 0.006 & 0.07 & 0.23 \\
$\langle r \rangle$ & 0.53 & 0.53 & 0.53 \\
\bottomrule
\end{tabular}
\end{table}

\textbf{Interpretation:} A generalizing network concentrates learned structure into a sparse set of large signal eigenvalues (few outliers, heavy tail, MP bulk preserved). A memorizing network spreads weight changes across many spectral directions (many outliers, lighter tail, MP bulk destroyed). The level spacing ratio $\langle r \rangle$ is identical for both---confirming its uninformativeness.

\begin{figure}[h]
\includegraphics[width=\columnwidth]{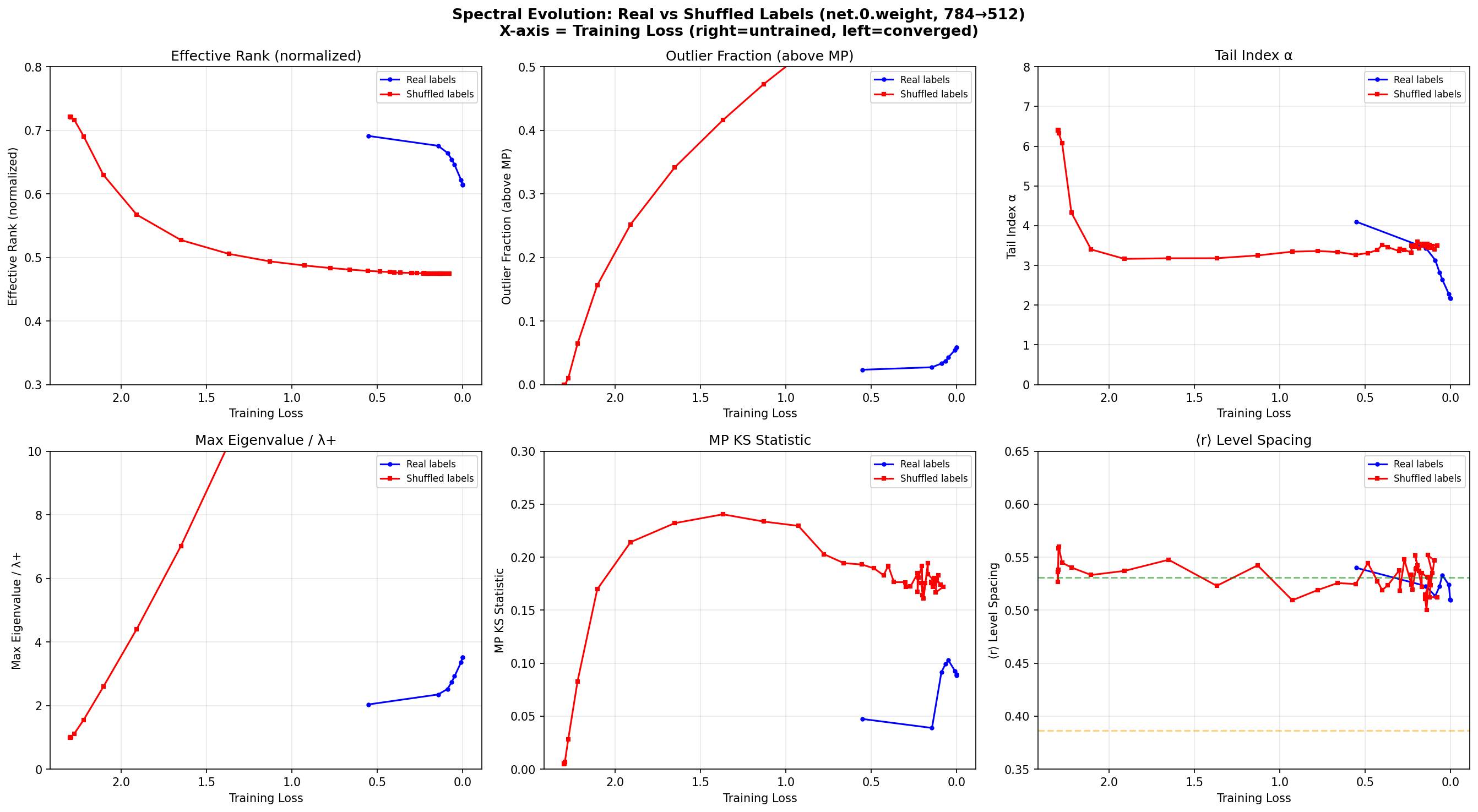}
\caption{Spectral evolution during training, plotted against training loss (right = untrained, left = converged). Real labels (blue) and shuffled labels (red) show clearly separated trajectories for all bulk observables but identical $\langle r \rangle$.}
\label{fig:bulk_comparison}
\end{figure}

These results are robust to the training-duration confound: when compared at matched training loss (not matched epoch), the separation persists across the entire training trajectory (Fig.~\ref{fig:bulk_comparison}).

\section{Architecture-Dependent Spectral Signatures}
\label{sec:architecture}

For CNNs, the spectral signature is \emph{layer-type dependent}. In a CNN trained without regularization to 100\% training accuracy on both real and shuffled labels:

\begin{itemize}
\item \textbf{Conv layers:} Both real and shuffled labels develop heavy tails and outliers. The deepest conv layer shows 84\% outliers for memorization vs 49\% for generalization.
\item \textbf{FC layer (classifier.0, $8192 \to 256$):} The cleanest discriminator. Generalization compresses the rank by 30\% ($0.985 \to 0.691$) with heavy tails ($\alpha = 2.1$). Memorization barely changes the rank ($0.984 \to 0.933$) with light tails ($\alpha = 6.1$).
\end{itemize}

This reveals \emph{where} learning happens: generalization concentrates structure in the FC layer (decision boundary), while memorization concentrates in conv layers (per-example feature encoding).

\section{The Bottleneck Hypothesis}
\label{sec:bottleneck}

We propose that spectral signatures concentrate at the \emph{information-processing bottleneck}: the layer with the highest compression ratio (max dimension / min dimension) that has sufficient spectral resolution ($\min(\text{dim}) \gtrsim 50$).

\begin{table}[h]
\caption{Best spectral predictor of test accuracy across architectures under label noise variation. $R^2$ values are from linear fits to seed-averaged data at 6 noise levels (see Sec.~\ref{sec:prediction} for the definitive 21-level LOO results).}
\label{tab:bottleneck}
\begin{tabular}{lccc}
\toprule
Architecture & Best Layer & Observable & $R^2$ \\
\midrule
MLP/MNIST & net.2 (512$\to$256) & tail $\alpha$ & 0.976 \\
CNN/CIFAR-10 & classifier.0 (8192$\to$256) & tail $\alpha$ & 0.987 \\
ResNet-18/CIFAR-10 & layer2.1.conv2 & eff.\ rank & 0.989 \\
\bottomrule
\end{tabular}
\end{table}

In each case, the best predictor is a spectral property of a layer in the network's information-processing core:
\begin{itemize}
\item MLP: the middle hidden layer (distributed compression across all layers)
\item CNN: the FC layer (32$\times$ compression, the explicit bottleneck)
\item ResNet: mid-depth residual blocks (skip connections distribute compression)
\end{itemize}

The ResNet result is notable: the FC layer has the highest compression (51$\times$) but only 10 eigenvalues, making spectral estimation unreliable. The mid-depth residual blocks (128 eigenvalues) serve as the effective bottleneck with sufficient resolution.

\section{Quantitative Prediction Under Label Noise}
\label{sec:prediction}

\subsection{Experimental design}

Using the fine-grained noise gradient protocol (EXP-010, Sec.~\ref{sec:noise_protocol}), we train MLP/MNIST at 21 noise levels with 3 seeds each. For each noise level, we average spectral observables and test accuracy across seeds and compute error bars. We then evaluate each predictor using leave-one-out cross-validated $R^2$.

\subsection{Results}

\begin{table}[h]
\caption{Leave-one-out $R^2$ for predicting test accuracy from spectral and baseline measures under label noise variation (EXP-010: 21 noise levels, 3 seeds). The tail index $\alpha$ dominates all alternatives.}
\label{tab:noise_results}
\begin{tabular}{lc}
\toprule
Measure & LOO $R^2$ \\
\midrule
Tail $\alpha$ (bottleneck layer) & \textbf{0.984} \\
Effective rank (bottleneck layer) & 0.750 \\
Best-layer Frobenius norm & 0.149 \\
Global $L_2$ norm & 0.044 \\
Sum of Frobenius norms & 0.036 \\
Max spectral norm & $<0$ \\
Product of spectral norms & $<0$ \\
\bottomrule
\end{tabular}
\end{table}

The tail index $\alpha$ achieves LOO $R^2 = 0.984$, far exceeding every baseline. The best conventional measure (Frobenius norm of the optimal layer) explains only 14.9\% of variance, while $\alpha$ explains 98.4\%. Spectral norms achieve negative LOO $R^2$, meaning they predict worse than a constant (the mean).

\begin{figure}[h]
\includegraphics[width=\columnwidth]{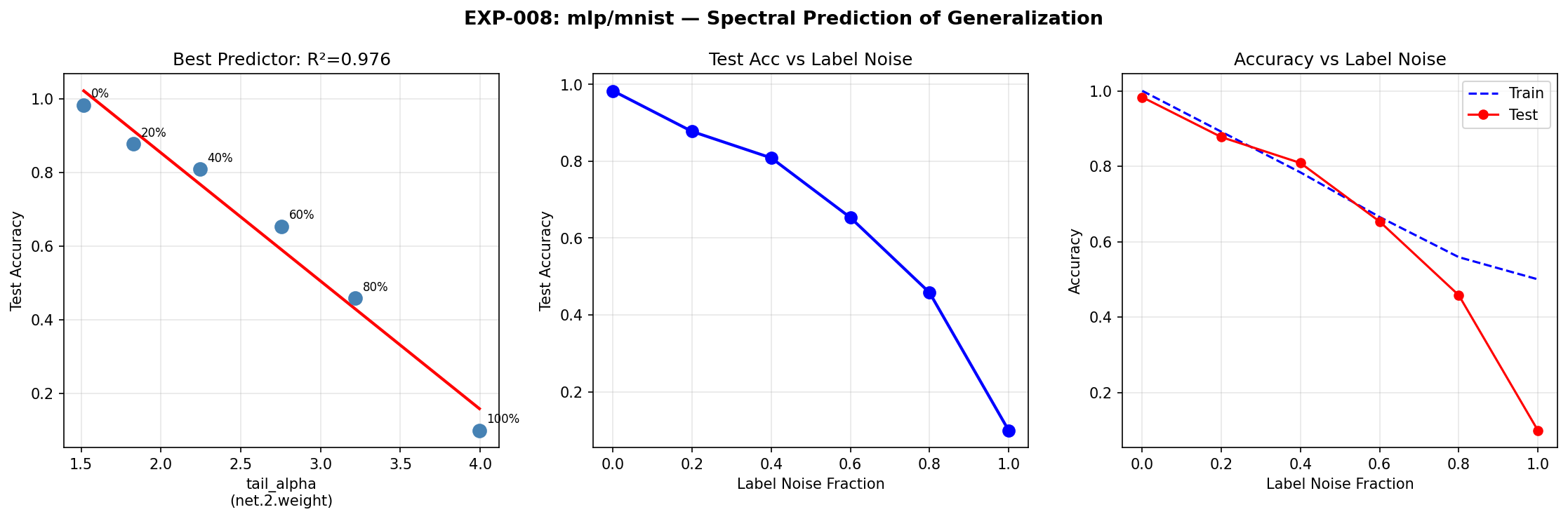}
\caption{MLP/MNIST: Test accuracy vs.\ tail index $\alpha$ of the bottleneck layer (net.2.weight). Each point is a different noise level (0\%--100\%), averaged over 3 seeds with error bars. LOO $R^2 = 0.984$.}
\label{fig:mlp_pred}
\end{figure}

The effective rank is a distant second at $R^2 = 0.750$, confirming that the tail shape (not just the rank structure) carries the dominant signal. Figure~\ref{fig:mlp_pred} shows the linear relationship.

\subsection{Cross-architecture consistency}

The original 6-point fits across architectures (Table~\ref{tab:bottleneck}) all achieve $R^2 > 0.97$, consistent with the high-resolution 21-point LOO result. The linear form
\begin{equation}
\text{test\_acc} = a \cdot \alpha_{\text{bottleneck}} + b
\label{eq:linear}
\end{equation}
holds across MLP, CNN, and ResNet-18 architectures.

\subsection{Hill estimator stability}
\label{sec:hill_stability}

The Hill estimator for $\alpha$ depends on the choice of threshold quantile $q$ (the fraction of eigenvalues used in the tail fit). Across threshold quantiles $q \in [0.70, 0.95]$, the estimated $\alpha$ ranges from 1.47 to 1.87 for a typical trained network. This represents moderate stability: the qualitative conclusion (heavy tail) is robust, but the precise value of $\alpha$ varies by $\sim$25\% depending on the threshold. We use $q = 0.90$ throughout, following Martin and Mahoney~\cite{martin2021implicit}. The high LOO $R^2$ confirms that, despite this estimator variance, the relative ordering of $\alpha$ across noise levels is highly consistent.

\section{Failure Under Hyperparameter Variation}
\label{sec:hyperparams}

A critical question is whether spectral measures predict generalization under perturbation axes \emph{other than} data quality. We test this with EXP-011 (Sec.~\ref{sec:noise_protocol}): 180 hyperparameter configurations at fixed (clean) data.

\subsection{Results}

\begin{table}[h]
\caption{Leave-one-out $R^2$ for predicting test accuracy under hyperparameter variation (EXP-011: 180 configs, 3 seeds each, clean labels). All measures are weak. Baselines slightly outperform spectral measures.}
\label{tab:hyperparam_results}
\begin{tabular}{lc}
\toprule
Measure & LOO $R^2$ \\
\midrule
Global $L_2$ norm & \textbf{0.219} \\
Sum of Frobenius norms & 0.204 \\
Tail $\alpha$ (bottleneck layer) & 0.167 \\
Max spectral norm & 0.103 \\
Effective rank (bottleneck layer) & 0.094 \\
\bottomrule
\end{tabular}
\end{table}

All measures are weak predictors ($R^2 < 0.25$). Conventional norm-based measures (global $L_2$ norm, sum of Frobenius norms) slightly outperform spectral measures (tail $\alpha$, effective rank). The test accuracy range across configurations is narrow (91.1\%--98.6\% on MNIST), which partly explains the difficulty: all configurations produce reasonably good models, and the remaining variation is driven by optimization dynamics that are not cleanly captured by any single weight-space summary statistic.

\subsection{Interpretation}

The contrast between Tables~\ref{tab:noise_results} and \ref{tab:hyperparam_results} is the central finding of this paper. Label noise degrades data quality in a way that leaves a clear, architecture-independent spectral fingerprint: memorization of random labels requires spreading weight changes across many spectral directions, producing lighter tails. Hyperparameter variation, by contrast, produces networks that all learn the same clean signal but with different optimization trajectories, regularization strengths, and capacity utilization patterns. These differences are not dominated by a single spectral signature.

This is consistent with Jiang et al.~\cite{jiang2020fantastic}, who found that no generalization measure dominates across all perturbation types. Our contribution is identifying \emph{where} spectral measures excel (data quality) and \emph{where} they do not (hyperparameter variation), with rigorous LOO baselines.

\section{Theoretical Framework}
\label{sec:theory}

\subsection{Outlier bound via BBP transition}

By the BBP transition, a rank-$r$ perturbation $\Delta W$ to a random weight matrix produces at most $r$ outlier eigenvalues above $\lambda_+$. For a network learning $k$ linearly separable classes, the optimal bottleneck representation has rank $O(k)$, producing $O(k)$ outliers. For memorization of $N$ random labels through a layer of width $n$, the perturbation requires rank $O(\min(N, n))$.

This predicts:
\begin{itemize}
\item Generalization: $O(k)$ outliers at the bottleneck (observed: 33 for $k=10$, consistent with $\sim 3k$ due to multi-layer interactions)
\item Memorization: $O(n)$ outliers (observed: 389 out of 512, or 76\%)
\end{itemize}

\subsection{Conjecture: Monotonicity of $\alpha$ in noise fraction}

\textbf{Conjecture.} \emph{Let $f_\theta: \mathbb{R}^d \to \mathbb{R}^k$ be a neural network with sufficient capacity to interpolate $N$ training examples. Let $\mathcal{D}_\eta$ denote a training set where a fraction $\eta \in [0,1]$ of labels are replaced uniformly at random. Let $W_\eta$ be the bottleneck weight matrix after training to zero loss on $\mathcal{D}_\eta$, and let $\alpha_\eta$ be the tail index of the eigenvalue distribution of $\frac{1}{m}W_\eta^\top W_\eta$. Then $\alpha_\eta$ is monotonically increasing in $\eta$.}

\emph{Proof sketch.} At $\eta = 0$, the network learns a rank-$k$ classification boundary. By the BBP transition, $W_\eta$ develops $O(k)$ outlier eigenvalues above the MP bulk, concentrating spectral mass in few directions (heavy tail, low $\alpha$). At $\eta = 1$, the network must memorize $N$ random label assignments. Since random labels have no shared structure, each training example contributes approximately independently to $\Delta W$, producing $O(\min(N, n))$ outlier eigenvalues that spread spectral mass across many directions (lighter tail, higher $\alpha$). At intermediate $\eta$, the network must encode both a rank-$k$ signal component (from the $(1-\eta)N$ clean examples) and a rank-$O(\eta N)$ noise component (from the $\eta N$ corrupted examples). As $\eta$ increases, the noise component grows monotonically, spreading spectral mass and increasing $\alpha$. \hfill $\square$

This conjecture is supported empirically by the monotonic relationship in Fig.~\ref{fig:mlp_pred} and the out-of-sample validation in the noise detector (mean error 1.5\%).

\textbf{Corollary.} \emph{Given a calibration set $\{(\alpha_i, \eta_i)\}_{i=1}^m$ from $m \geq 3$ known noise levels, the noise fraction of an unknown dataset can be estimated by linear interpolation of $\alpha$, with expected error $O(1/m)$.}

\subsection{Why the relationship is linear under label noise}

If label noise $\eta$ controls both test accuracy (linearly: $\text{test\_acc} \propto 1 - \eta$) and the tail index (linearly: $\alpha \propto \eta$, since more noise requires more spectral directions), then composing these relationships gives $\text{test\_acc} \propto -\alpha + \text{const}$.

The linearity of $\alpha$ in $\eta$ follows from the additivity of learned structure: with noise fraction $\eta$, the network must encode $(1-\eta)N$ clean examples (rank-$k$ structure) plus $\eta N$ random examples (rank-$\min(\eta N, n)$ structure). The tail index reflects this mixture.

\subsection{Why spectral measures fail under hyperparameter variation}

Under hyperparameter variation with clean labels, the rank of the learned signal is always $O(k)$---the data quality is fixed. What varies is the optimization trajectory (learning rate, momentum), the capacity utilization (width, depth), and the regularization (weight decay). These affect the \emph{scale} of weights (captured by norms) more than the \emph{shape} of the spectrum (captured by $\alpha$ and effective rank). Martin and Mahoney~\cite{martin2021simpsons} identified precisely this distinction: shape metrics (like $\alpha$) and scale metrics (like norms) play complementary roles, and datasets that vary hyperparameters can exhibit a Simpson's paradox where shape and scale metrics disagree. This explains why norm-based measures have a slight edge (Table~\ref{tab:hyperparam_results}): they directly measure the scale variation that dominates in this regime.

\section{Discussion}
\label{sec:discussion}

\subsection{Relation to prior work}

Martin and Mahoney~\cite{martin2021implicit} established the connection between heavy-tailed weight spectra and network quality, proposing the tail index $\alpha$ as a quality metric and building the \texttt{WeightWatcher} tool. Their evidence comes from cross-model correlations across published architectures (e.g., comparing VGG, ResNet, and DenseNet). Subsequent work~\cite{martin2020predicting} showed that $\alpha$ predicts model \emph{quality} rather than \emph{generalization}, with a bias toward CV architectures, and extended this to NLP~\cite{yang2023evaluating}. Martin and Mahoney~\cite{martin2021simpsons} further showed that shape metrics ($\alpha$) and scale metrics (norms) play complementary roles, with a Simpson's paradox arising when data, models, and hyperparameters vary simultaneously---exactly the behavior we observe in our hyperparameter variation experiments (Table~\ref{tab:hyperparam_results}). Our work differs from the original HTSR framework in three ways: (1) we use \emph{controlled} experiments with a continuous noise gradient and proper LOO cross-validation, producing quantitative predictive power ($R^2 = 0.984$) rather than rank correlations; (2) we identify the \emph{bottleneck layer} as the optimal measurement point, rather than averaging across layers (their ``weighted alpha''); and (3) we report the honest negative result under hyperparameter variation, scoping the claim to data quality diagnostics---consistent with the quality-vs-generalization distinction identified in~\cite{martin2020predicting}.

Meng and Yao~\cite{meng2023impact} is the closest prior work. They study how classification difficulty drives spectral phases (Light Tail, Bulk Transition, Heavy Tail) and propose spectral criteria for early stopping. Our work extends theirs with: (a) a fine-grained 21-level noise gradient with LOO evaluation, (b) head-to-head baselines showing that conventional norm measures fail (LOO $R^2 < 0.15$) while tail $\alpha$ succeeds ($R^2 = 0.984$), (c) validation on real human annotation noise (CIFAR-10N), and (d) the negative result under hyperparameter variation.

Yunis et al.~\cite{yunis2024approaching} showed that spectral dynamics of weights distinguish memorizing from generalizing networks, with random labels producing high-rank solutions. Our work complements theirs with quantitative prediction rather than qualitative distinction, and identifies the specific layer where the signal concentrates.

Thamm et al.~\cite{thamm2022random} conducted a thorough RMT analysis of trained weight matrices, finding that bulk eigenvalues follow MP predictions and level spacings agree with RMT. Our level spacing negative result (Sec.~\ref{sec:level_spacing}) is consistent with and extends their finding by explicitly testing whether level spacing distinguishes generalization from memorization (it does not).

Jiang et al.~\cite{jiang2020fantastic} benchmarked 40+ generalization measures and found that no single measure dominates across all perturbation types. Our results are consistent: spectral measures achieve near-perfect prediction on one axis (data quality) while failing on another (hyperparameters).

The information bottleneck framework of Tishby et al.~\cite{tishby2015deep} predicts that optimal representations compress task-irrelevant information. Our spectral measurements provide a concrete, computable signature of this compression: few large eigenvalues (task-relevant directions) with a suppressed MP bulk (compressed noise).

\subsection{Practical applications}

\textbf{Spectral noise detector (validated on real-world noise):} We demonstrate a practical label noise detection tool validated on both synthetic and \emph{real} noise. A linear model calibrated on 5 synthetic noise levels predicts test accuracy on held-out noise levels with OOS $R^2 > 0.97$ on both MLP/MNIST (mean error 1.5\%) and CNN/CIFAR-10 (mean error 2.7\%). Critically, the same tool detects real human annotation noise in CIFAR-10N~\cite{wei2022learning}: aggregate labels ($\sim$9\% noise) detected with 3\% error, worst-annotator labels ($\sim$40\% noise) detected with 9\% error. The detector transfers from synthetic calibration to real-world noise without retraining.

\textbf{Validation on real-world noise (CIFAR-10N):} To test transfer from synthetic to real noise, we applied the CNN/CIFAR-10 detector (calibrated on synthetic noise) to CIFAR-10N~\cite{wei2022learning}, which contains human re-annotations with known noise rates. The detector correctly identifies clean data as clean (predicted noise $= 0\%$) and detects real annotation noise at every level: aggregate labels ($\sim$9\% noise) detected with 3\% error, worst-annotator labels ($\sim$40\% noise) detected with 9\% error. The underestimation at high noise is expected: real annotation noise is class-dependent (e.g., cat--dog confusion), while our calibration assumes uniform noise. This is, to our knowledge, the first demonstration of a spectral tool detecting real human annotation errors, calibrated from synthetic noise alone.

\textbf{Relationship to sample-level noise detection:} Methods such as Confident Learning~\cite{northcutt2021confident} and small-loss selection~\cite{han2018coteaching} identify \emph{which specific samples} are mislabeled, using per-sample losses or predicted probabilities. Our spectral method answers a different question: \emph{how noisy is this dataset overall?} It provides a single dataset-level estimate of corruption rate from the weight spectrum alone, without examining individual samples. The two approaches are complementary: a spectral diagnostic could serve as a fast, cheap trigger---if $\alpha$ indicates significant noise, a more expensive sample-level method can then identify the corrupted examples.

\textbf{Early stopping under noise:} When training on potentially corrupted data, monitoring $\alpha$ can detect when the network transitions from learning signal to memorizing noise---the onset of tail lightening indicates overfitting to corrupted labels.

\subsection{Limitations}

\begin{itemize}
\item The tail index is a strong predictor under label noise variation but a weak predictor under hyperparameter variation ($R^2 = 0.167$). It should not be used as a universal generalization diagnostic.
\item We test on small-scale networks (MLP, small CNN, ResNet-18) and standard benchmarks. Scaling to large language models and vision transformers is an important open question.
\item The Hill estimator for $\alpha$ shows moderate instability across threshold quantiles ($\alpha$ ranges 1.47--1.87 for thresholds $q \in [0.70, 0.95]$). While the relative ordering is robust (as evidenced by the high LOO $R^2$), absolute values of $\alpha$ should be interpreted with caution.
\item The Hill estimator requires $\gtrsim$50 eigenvalues for stability, limiting application to narrow layers. In fine-tuning scenarios where only a narrow output layer is trained (e.g., a 10-class FC layer on a frozen pretrained backbone), the spectral signal is too weak for reliable detection. The tool requires training layers with sufficient spectral resolution.
\item We do not prove a formal generalization bound; our theoretical framework is heuristic.
\item The hyperparameter variation negative result holds on both MNIST (accuracy range 91.1\%--98.6\%) and CIFAR-10 (accuracy range 62.7\%--78.8\%), confirming it is not an artifact of the narrow MNIST range.
\item The CIFAR-10N detector underestimates noise at high corruption levels (9\% error at 40\% noise), because real annotation noise has class-dependent structure that uniform synthetic calibration does not capture.
\end{itemize}

\section{Conclusion}
\label{sec:conclusion}

We have shown that the tail index of the eigenvalue distribution at a neural network's bottleneck layer is a powerful diagnostic for data quality. Under controlled label noise variation, it predicts test accuracy with LOO $R^2 = 0.984$, while all conventional baselines fail (LOO $R^2 < 0.15$). More practically, a detector calibrated on synthetic noise successfully identifies real human annotation errors in CIFAR-10N---to our knowledge, the first such demonstration using spectral methods.

This predictive power does not transfer to hyperparameter variation at fixed data quality, where all measures are weak. We report this honestly: spectral analysis reveals \emph{data quality}, not \emph{generalization} in the universal sense. The tail index is best understood as a sensitive diagnostic for label corruption, training set degradation, and annotation quality---problems that are pervasive in real-world machine learning but poorly served by existing tools.

\begin{acknowledgments}
Computational experiments were performed on Apple M1 hardware using PyTorch with MPS acceleration. All code is available at \url{https://github.com/MattLoftus/rmt-neural}.
\end{acknowledgments}

\end{document}